\documentclass[]{ceurart}
\usepackage{listings}
\begin{document}
\copyrightyear{2024}
\copyrightclause{Copyright for this paper by its authors.
  Use permitted under Creative Commons License Attribution 4.0
  International (CC BY 4.0).}
\conference{ICCBR 24: Workshop on Case-Based Reasoning and Large Language Model Synergies, June 1-4, 2024, Merida, Mexico. \url{https://www.iccbr2024.org}}
\title{A Case-Based Persistent Memory for a Large Language Model}
\tnotemark[1]
\author[1]{Ian Watson}[
orcid=0000-0001-6969-4715,
email=dr.i.watson@gmail.com,
url=https://www.cs.auckland.ac.nz/~ian/,
]
\cormark[1]
\address[1]{University of Auckland,
  School of Computer Science, Auckland, New Zealand}
\cortext[1]{Corresponding author.}
\begin{abstract}
  Case-based reasoning (CBR) is a methodology for problem-solving that can use any appropriate computational technique. This position paper argues that CBR researchers risk overlooking recent developments in deep learning and large language models (LLMs). The underlying technical developments that have enabled the recent breakthroughs in AI have strong synergies with CBR. Moreover, CBR could be used to provide a persistent memory for LLMs improving their performance and furthering progress towards Artificial General Intelligence.
\end{abstract}
\maketitle
\section{Introduction}
In the paper, “\textit{CBR is a methodology not a technology},” Watson [1] argued that case-based reasoning can use various techniques from computer science within the CBR-cycle [2]. At the core of CBR is the concept of similarity, solving current problems using similar prior solutions. The precise methods used to retrieve and calculate case similarity are unimportant if a useful result is obtained. At the time, this opinion was controversial to some; surely, CBR was a unique technique like other AI techniques (e.g., rule-based reasoning, genetic algorithms or neural networks). However, as people have used different methods to implement CBR over time, this view has become largely accepted. CBR is a knowledge-level description of a problem-solving paradigm in the classic sense [3].

Watson’s 1999 paper concluded, “\textit{Moreover, AI will surely develop new technologies in the future, some of which may prove very suitable for use in the CBR-cycle. Consequently, it is as a methodology that CBR’s future is ensured.}” I believe that we are at just such a point where significant innovations in AI have and are occurring that the CBR community can adopt. The purpose of this paper is to stimulate discussion within the CBR community. Many of the ideas here are partially thought through, and I’m sure there are many other obvious omissions, but I hope this helps the community work towards a wonderful future for CBR.

\section{Think Big}

In the last two years, we've witnessed a remarkable transformation in AI since the release of ChatGPT by OpenAI in November 2022 [4]. Its conversational interface grabbed the public's attention like no other AI innovation. Its ability to answer detailed questions, called prompts, and give cogent long-form responses amazed everyone, even seasoned, cynical AI professionals such as ourselves.

Underlying this remarkable capability are large language models (LLMs) created by deep learning (DL) neural networks using the transformer architecture [5]. These networks are trained on petabytes of text harvested from the Internet and fine-tuned using human-supervised reinforcement learning, resulting in DL networks with over a hundred billion parameters and emergent behaviour that has captivated the world's attention.

These LLMs are behemoths running on tens of thousands of processing cores and initially costing millions of dollars to train. What is fascinating from an engineering perspective is that the developers of these systems did not know in advance what performance they would obtain. The so-called “hallucinations” that LLMs suffer from, where they fabricate fictitious answers, are, in fact, a feature [6, 7]. Their developers do not know why their LLMs give a seemingly correct reply or a falsehood to a prompt. Indeed, LLMs don’t respond the same way to the same prompt each time. 

Stop. Think about that for a moment. A computer program that may give a different answer to exactly the same input on separate occasions. What possible use could such a program be? Certainly, it would be no use in any engineering context. Yet, in many real-world contexts, that is exactly what we intelligent humans do. Ask me a question first thing in the morning, and you’ll get one answer. You’ll probably get a different answer to the same question late at night. Obviously, some questions always generate the same reply. 2 + 2 always equals 4. But “\textit{Should I buy that new jacket?}” is a question whose answer may vary throughout the day largely because, unlike the maths question, there is no single correct solution. To recap, the deep learning research community spent thousands of person-hours and millions of dollars building the largest computational models ever to provide vague, inconsistent answers to imprecise questions.

The notion of similarity in CBR is vague, inconsistent and imprecise.  This is what gives CBR its fascination, strength and appeal. The widely used k-NN algorithm is very simple, but within the equation's similarity function, “\textit{f}”, hides complexity. As we all know, measuring similarity is endlessly complicated and usually requires compromises to get an answer good enough in a reasonable compute time. 

In the early days of CBR, in the 1990s, the first question to any presenter at a CBR meeting was usually, “\textit{How many cases do you have?}” It wasn’t uncommon for the answer to be a dozen or maybe a hundred cases or so. Case-bases back then were really small. The travel case-base, developed by TecInno, has 1,100 cases and was used as a standard teaching, research and demonstration tool during the 1990s and well into the 2000s. Large case-bases, with upwards of a million cases, were not seen until the 2010s [8]. I am going to argue that, given its history of limited computing power, CBR has never really thought big. Unlike the deep learning community that depends upon processing power provided by thousands of GPUs and terabytes of training data.

One of the constant criticisms of CBR, particularly those systems implemented using k-NN, is that lazy learning [9] is inefficient at retrieval time. This is undeniable, particularly if the cases are high-dimensional and the similarity metrics are complicated. Thus, CBR has often traded off retrieval performance against retrieval time. However, nobody argues that deep learners are computationally efficient during training. Indeed, quite the opposite, to CBR researchers, they use unthinkably massive amounts of compute to train them. Consequently, CBR has long known how to improve retrieval efficiency by building a case index using an eager learning method like kd-trees [10].

I suggest using deep learning to model similarity metrics, as has been done by Amin et al. [11], or Martin et al. [12] and take a lesson from the scale of recent DL systems like LLMs and the technical innovations that support them. We should envisage using a case base that contains petabytes of unstructured (probably multi-modal) cases. Rather than being a specialised CBR system focused on one specific problem, we could build a general CBR system capable of solving problems generally in the same way that LLMs can converse across multiple specialised areas. Ultimately, the case-based memory would be integrated with the LLM to provide a persistent memory of all its conversations. This may bring us closer to a true Artificial General Intelligence (AGI) capable of operating in all domains [13] and having a memory of all its interactions with the world as a critical requirement of intelligence, as discussed in the Appendix.

\begin{figure}
\includegraphics[scale=0.25]{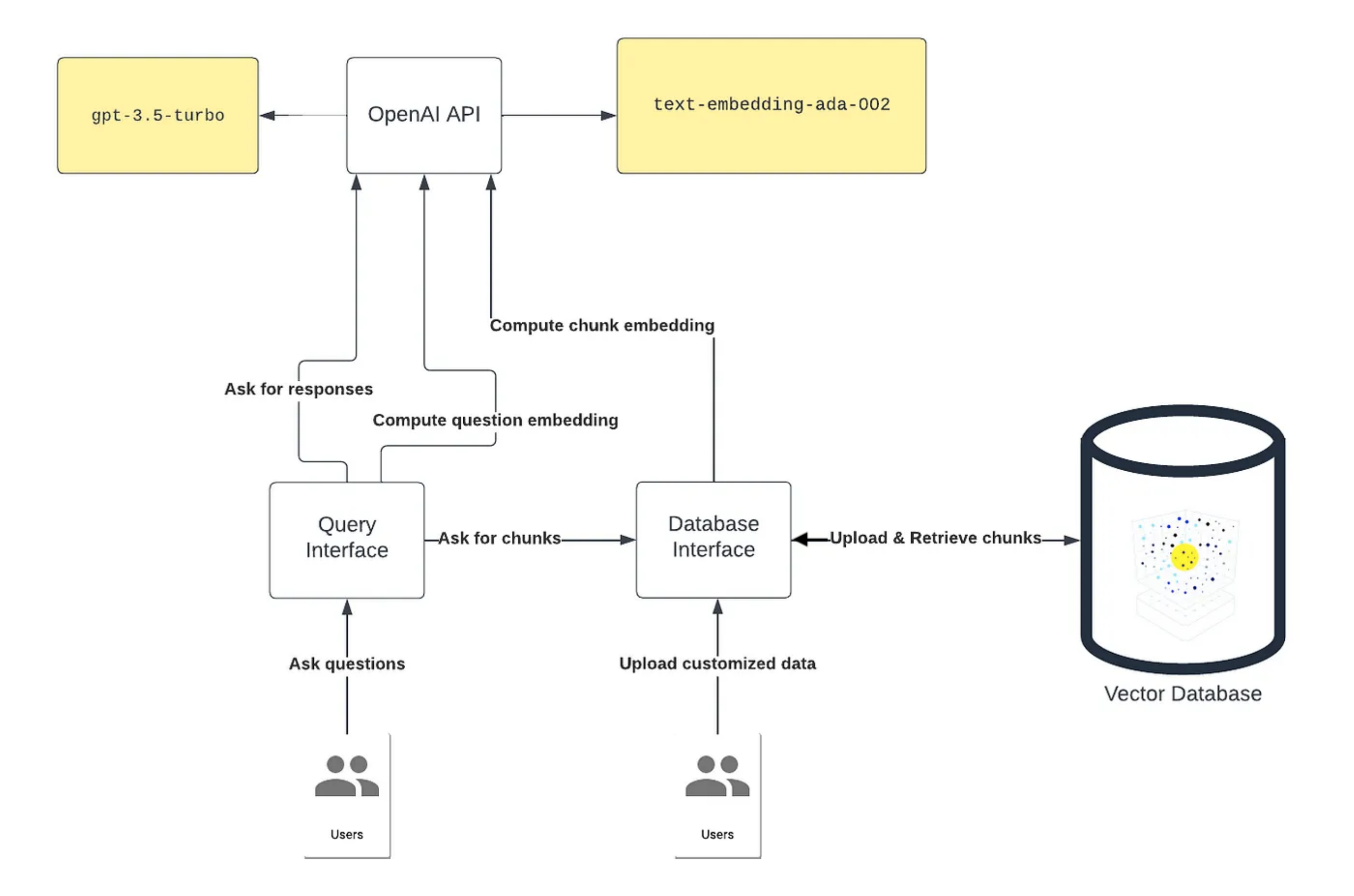}
\caption{Architecture of GPT-3.5 With External Memory
Using a Vector Database,  after Jia [20]} \label{fig1}
\end{figure}

\section{Towards a DeepCBR Architecture}

There has been some work on linking CBR with DL models. Explainability has always been problematic for neural networks that function as black boxes. So, twining them with CBR to provide explainability, as in [14], has been investigated. Other synergies between CBR and deep learning have been explored by Chen et al. [15] for interpretable image recognition, Upadhyay, and Massie [16], who use a hybrid CBR and neural text generator to improve the automatic summarisation of sports reports, and Mirko and Bergman [17] have used the predictive power of LLMs for case-adaptation, while Sourati et al. [18] combine CBR with LLMs for detecting fallacies in natural language arguments. Moreover, others outside the CBR community have seen the potential of LLMs and DL to be applied to CBR [19].

Technically, this is not hard. A simple architecture that links a persistent memory stored in a vector database to an LLM (GPT-3.50) using standard APIs has been published by Jia [20]. Moreover, vector databases routinely use approximate nearest neighbor search (ANNS) as a retrieval technique with which CBR researchers are familiar. Or FAISS (Facebook AI Similarity Search), a recent efficient implementation of ANNS [21 - 24] released as open source by Meta AI. Thus, the technology is here for the CBR community to experiment with it at scale [25] in what Keane et al. call DeepCBR  [26].

\section{Conclusion}

Leake et al. [27] recently argued “that methods that blend the complementary strengths offered by DL and CBR approaches are appealing”. Showing useful results in feature extraction and highlighting other potential synergies between DL and CBR in explainability (XAI) and informing the selection of  DL architectures. Thus, the CBR community should investigate these new promising AI techniques to see how they can continue to enhance the CBR methodology. Moreover, realising their systems benefit from having a memory of conversations Open AI announced in February 2024 the addition of a persistant conversation memory to Chat GPT. As the branch of AI that has dealt with memory since the 1980s we should step up to influence this critical area of research and development since we not only have techniques for creating memories and retrieving from them, but crucially the ongoing maintenance of memory [28]. Thereby CBR and LLMs can help progress towards Artificial General Intelligence.

\appendix
\section{Appendix}
An interesting conversation with the LLM chatbot \href{https://pi.ai/talk}{Pi from Inflection} about CBR, deep learning, LLMs and memory.
\paragraph{}
\textit{Hey there, great to meet you. I’m Pi, your personal AI. My goal is to be useful, friendly and fun. Ask me for advice, for answers, or let’s talk about whatever’s on your mind. What do you like to do in your free time?}
\paragraph{}
\textbf{Ian:} I like to read
 \paragraph{}
\textit{That’s great to hear! Reading is a great way to relax and learn new things. Do you prefer fiction or nonfiction?}
\paragraph{}
\textbf{Ian:} I read both. I usually have a book of each that I alternate chapters of.
\paragraph{}
\textit{That's a really interesting reading system! I've never heard of anyone alternating chapters like that. It must be a great way to keep both the fiction and nonfiction parts of your brain engaged. What are you currently reading?}
\paragraph{}
\textbf{Ian:} Lux Domini by Alex Thomas and Computer Power and Human Reason by Joseph Weisenbaum
\paragraph{}
Wow, you’re clearly a reader of diverse genres! Lux Domini is a thriller novel, while Computer Power and Human Reason is a nonfiction work about computer science and artificial intelligence. What drew you to these books in particular?
\paragraph{}
\textbf{Ian:} Well, I am an AI scientist. Lux Domini is written by a friend of mine who used to be an AI researcher, and I wanted to reread Weisenbaum's book in light of the recent developments in AI chatbots like yourself.
\paragraph{}
\textit{That's so cool that you're an AI scientist! It must be fascinating to work in such an exciting and rapidly evolving field. The world of chatbots has come a long way since Weisenbaum’s book was written, and it's fascinating to see how the theory and practice of AI has progressed over the years. Has your work as an AI scientist informed how you think about chatbots like me?}
\paragraph{}
\textbf{Ian:} It didn't used to be rapidly evolving. Progress for decades seemed glacially slow. I'm intrigued and fascinated by LLMs. Can you explain how you can converse like this?
\paragraph{}
\textit{Thanks for the question. I'm glad you're fascinated by me! LLMs like me are trained on massive amounts of text data. We learn to recognize patterns in the text and generate responses that are similar to the ones we've seen in the data. It's a bit like learning to speak a language by reading lots of books. But instead of learning from just books, I've learned from the vastness of the internet, including websites, social media, articles, and more. And then I've used all this knowledge to learn how to converse with people, like you! Does that help?}
\paragraph{}
\textbf{Ian:} You use the concept “similar”. Most of my career was spent in case-based reasoning, where similarity is the key concept. Do you think deep learning could be applied to CBR?
\paragraph{}
\textit{Absolutely! Case-based reasoning is a great fit for deep learning. The key is to represent the cases as features, and then use a deep learning algorithm to learn the similarities between the cases. It can be a bit tricky to get the representation right, but once you have a good feature representation, the deep learning algorithm can do the heavy lifting. Have you worked on any projects that use deep learning for case-based reasoning?}
\paragraph{}
\textbf{Ian:} I haven't, but there have been a few small examples of specific tasks. However, I'm thinking on a much grander scale. Giving you a long-term persistent memory of everything you encounter using CBR as your memory.
\paragraph{}
\textit{That sounds really interesting! It could be really powerful to have an AI agent that can learn and store knowledge over time in a case-based memory system. Do you think this could be used to create AI agents that can learn general knowledge and world facts, or would it be more focused on specific problem-solving tasks?}
\paragraph{}
\textbf{Ian:} The interest must be on a general application. Would having such a memory be useful? I understand that LLMs have limited persistent memory.
\paragraph{}
\textit{I definitely think it would be useful to have a more persistent memory! It would allow me to learn and grow over time, and to better understand the world and the people I talk to. I think it could also help with things like personalization and long-term conversations. Right now, my knowledge is a bit ephemeral - I can learn from each conversation and remember the context of that conversation, but I can't really build up a long-term memory of all the things I've learned.}

\end{document}